\documentclass[conference]{IEEEtran}

\usepackage[english]{babel}
\usepackage[utf8x]{inputenc}
\usepackage{amsmath}
\usepackage{graphicx}
\usepackage{cite}

\title{
BuildSeg: A General Framework for the Segmentation of Buildings
}

\author{
Lei Li$^1$, Tianfang Zhang$^2$, Stefan Oehmcke$^1$, Fabian Gieseke$^3$, Christian Igel$^1$ \\

$^1$Department of Computer Science, University of Copenhagen, Copenhagen, Denmark \\
$^2$School of Information and Communication Engineering, University of Electronic Science and Technology of China, China \\
$^3$Department of Information Systems, University of Münster, Münster, Germany

}

\begin{document}

\maketitle
\begin{abstract}
Building segmentation from aerial images and 3D laser scanning (LiDAR) is a challenging task due to the diversity of backgrounds, building textures, and image quality. 
While current research using different types of convolutional and transformer networks has considerably improved the performance on this task, even more accurate segmentation methods for buildings are desirable for  applications such as automatic mapping.
In this study, we propose a general framework termed  \emph{BuildSeg} employing a generic approach that can be quickly applied to segment buildings. 
Different data sources were combined to increase generalization performance.

The approach yields good results for different data sources as shown 
by experiments on high-resolution multi-spectral and LiDAR imagery of cities in Norway, Denmark and France.

We applied ConvNeXt and SegFormer based models on the  high resolution aerial image dataset from the MapAI-competition.
The methods achieved an IOU of {0.7902}  and a boundary IOU of {0.6185}.  
We used post-processing to account for the rectangular shape of the objects. 
This increased the boundary IOU from {0.6185} to {0.6189}.

\end{abstract}

\section{Introduction}
Detecting buildings from remote sensing imagery has been extensively studied~\cite{liow1990use, li2008adaptive, ferraioli2009multichannel} as it is of great importance for many fields, such as urban planning, population estimation, economic development, and topographic map production.
Since the amount of data cannot be processed manually, data-driven machine learning methods are needed to reduce the manual work required to obtain reliable urban development mappings.

Segmenting buildings on a large scale is a challenging task because satellite or aerial images can be very diverse, for example, because of different styles of architecture, building materials, and topography. 
Quite a number of benchmarks for the segmentation of buildings have been published~\cite{maggiori2017dataset, MnihThesis, Jyhne2022, 9460988, roscher2020semcity, ji2018fully, rottensteiner2013isprs, li2021pointflow, weber2021artifive}. 
Since the silhouettes of buildings can be very different, combining several datasets with different characteristics can lead to more generally applicable building segmentation models. 

In this study, we propose a framework
for building segmentaion referred to as \emph{BuildSeg}.
We consider the Inria Aerial Image Labeling Benchmark~\cite{maggiori2017dataset} combined with the MapAI-competition dataset~\cite{Jyhne2022} to improve the segmentation performance of the latter. 
When designing the BuildSeg framework, our goal was to design a segmentation pipeline that is generally applicable. 
Therefore, several benchmarks and corresponding models are available within the framework~\cite{maggiori2017dataset, MnihThesis, Jyhne2022, 9460988, roscher2020semcity, ji2018fully, rottensteiner2013isprs, li2021pointflow, weber2021artifive}.

\begin{figure}
    \centering
    \includegraphics[scale=0.16]{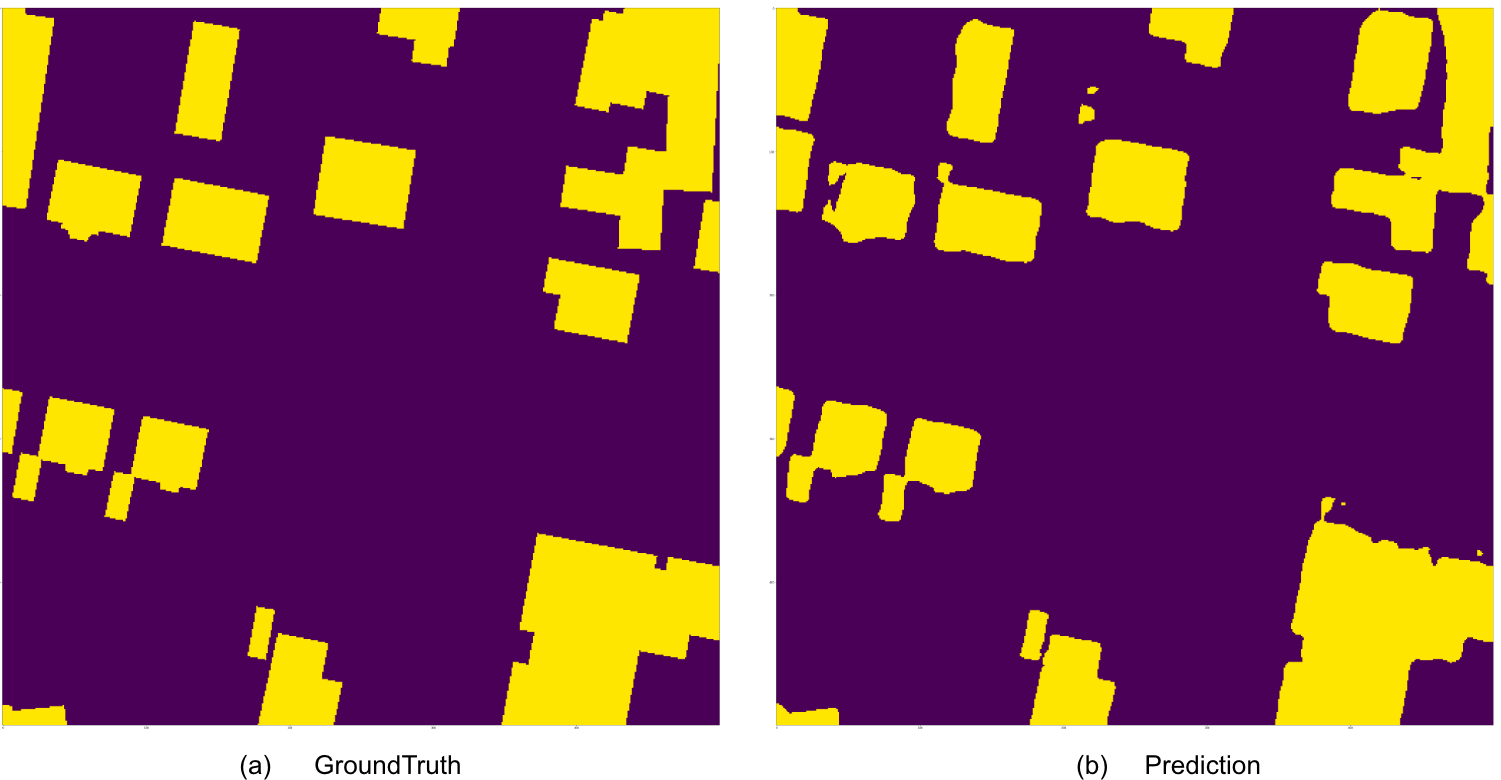}
    \caption{Examples from the MapAI-competition~\cite{Jyhne2022}. 
    The left image~(a) shows the ground-truth data, and the right image~(b)  the prediction created by the ConvNeXt model.}
    \label{fig:results}
\end{figure}

Neural networks and in particular convolutional neural networks (CNNs) have become the go-to methods for image segmentation, see 
\cite{wang2022comprehensive} for a recent review.
The U-Net~\cite{ronneberger2015u} is one of the fundamental segmentation architectures, and we have been successfully applying it to  remote sensing imagery (e.g., \cite{brandt:20,hellweg:22}).
It uses an encoder/decoder structure which processes the input image at different scales and allows to detect high-frequency patterns while being computationally feasible.

The original U-Net architecture can be generalized by replacing 
the encoder and decoder by tailored networks. This makes the U-Net very versatile and allows to utilize state-of-the-art encoders.
In our framework, we consider two different U-Net variants, SegFormer~\cite{xie2021segformer} and ConvNeXt U-Net~\cite{ConvNeXt}, where the decoder of the ConvNeXt U-Net are backwards strided convolutions~\cite{long2015fully}.

The main contributions of this study can be summarized as follows:
(1) we propose a general framework called BuildSeg based on~\cite{mmseg2020} for 
segmenting buildings in aerial images of different resolutions; 
(2) we explore how 3D information from LiDAR affects the performance of deep CNN models; 
(3) we combine different datasets and apply rectangle-aware post-processing to create rectangular boundaries that match the labels more accurately. 
The proposed approach achieved the IOU of {0.7902} for the segmentation of images in \emph{MapAI: Precision in Building Segmentation}~\cite{Jyhne2022} benchmark.
\begin{figure}
    \centering
    \includegraphics[scale=0.15]{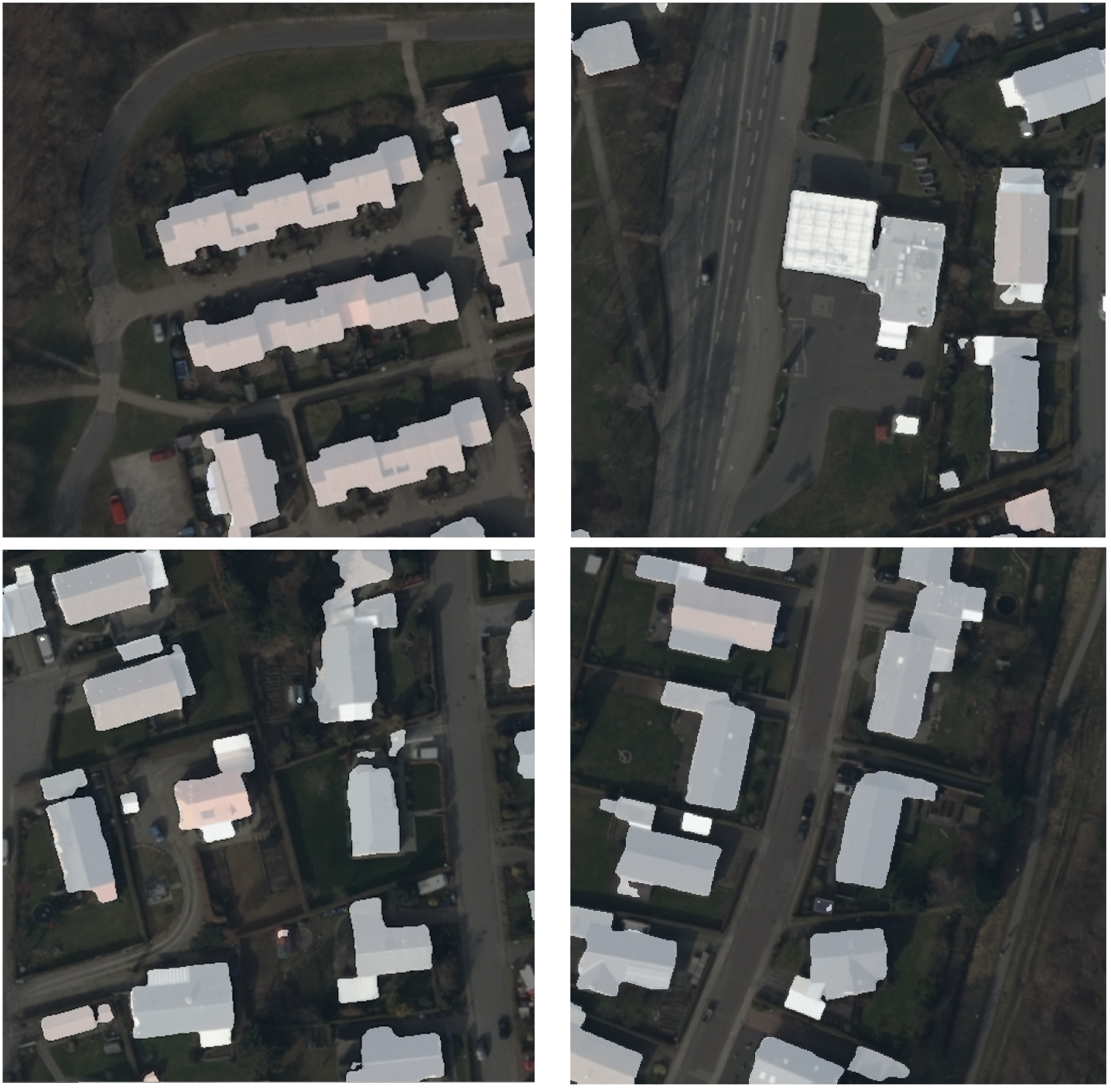}
    \caption{Building predictions with SegFormer, overlaid are the prediction result of the model.}
    \label{fig:prediction}
\end{figure}

\section{Method}
We developed our framework for the MapAI challenge~\cite{Jyhne2022}, which provides both aerial images and LiDAR data. 
The challenge formulates two tasks. 
The first is the segmentation of buildings only using the aerial imagery. 
In the second task, the LiDAR data must be segmented either with or without  aerial images. 

We additionally used the data from~\cite{maggiori2017dataset} to improve the performance.
A subset of {5000} images was considered  as additional training data. 
To align the image sizes, we cropped the input images to \(500\times500\).

To further increase diversity in the data, we applied augmentations including random cropping, random vertical and horizontal axis flipping, and random changes to brightness ($\delta=32$), contrast (range: $0.5 - 1.5$), saturation (range: $0.5 - 1.5$), and hue ($\delta=18$). 

We tried different models such as the standard U-Net and variants of it, namely ConvNeXt and SegFormer~\cite{xie2021segformer}. 
For the ConvNeXt model, ConvNeXt~\cite{ConvNeXt} is used as encoder and backwards strided convolution~\cite{long2015fully} as  decoder. 
We also tried EfficientNet~\cite{pmlr-v97-tan19a} as encoder but the results were not as good.
All encoders were pre-trained on ImageNet.

Two  metrics were considered  to measure the performance: intersection over union (IOU) and boundary intersection over union (BIOU)~\cite{cheng2021boundary}. 

LiDAR height data were added directly as an additional channel to the multi-spectral data when available. 

For post-processing, we applied a sequence of morphological opening and closing operations to detect lines and then removed points not matching the hypothesis of a rectangular structure.

\section{Experimental Results}
The results are summarized in  Table~\ref{table:exp}. 
The model SegFormer-B5 performed best  in terms of IOU and BIOU for the images. 
Note that SegFormer-B5 and SegFormer-B4 have more layers than SegFormer-B0.

\begin{table}
\caption{Performance of different models on the MapAI-competition image test set (without post-processing). As baseline we show a standard U-Net~\cite{ronneberger2015u}.}
 \label{table:exp}
\begin{center}

    \begin{tabular}{@{}l|c c c@{}}

    Model & {IOU} & {BIOU} \\
        U-Net & 0.7611 & 0.5823 \\
        ConvNext &  0.7841 & 0.6105 \\
        SegFormer-B0 & 0.7632 & 0.5901 \\
        SegFormer-B4 & 0.7844 & 0.6116 \\
        SegFormer-B5 & 0.7902 &  0.6185   \\
    \end{tabular}
\end{center}
\end{table}

Figure~\ref{fig:prediction} illustrates results of the SegFormer-B5 model, and it can be seen that the buildings were nicely captured. 

The averaged score, computed as the mean of IOU and BIOU,  was {0.7044} without post-processing and slightly increased to  {0.7045} after post-processing, so the latter should be preferred if IOU and BIOU weight the same.
When combined with LiDAR, the method reached an IOU of {0.8506} and a BIOU of {0.7461}.

\section{Conclusion}
For the MapAI-competition, we proposed a solution that utilizes additional building datasets and current state-of-the-art deep learning architectures.
The method achieved an IOU of {0.7902} and boundary IOU of {0.6185} for  the task of segmenting buildings in aerial images. 
Using additional information from LiDAR further improved the results, increasing the IOU to {0.8506} and the BIOU to {0.7461}.


\newpage
\bibliographystyle{IEEEtran}
\bibliography{ref}

\end{document}